%% file: root.tex
\documentclass[letterpaper, 10 pt, conference]{ieeeconf}  

\IEEEoverridecommandlockouts                              

\usepackage{graphics} 
\usepackage{epsfig} 
\usepackage{times} 
\usepackage{amsmath,amssymb,amsfonts}
\usepackage{multirow}
\usepackage{tabularx}
\usepackage{balance}

\usepackage{calc}

\usepackage{caption}
\DeclareCaptionLabelSeparator{dottilde}{.~~}
\usepackage{hyperref}
\usepackage{cleveref}
\usepackage{booktabs}

\usepackage{color} 
\usepackage{soul} 

\title{\LARGE \bf
Optimizing NeRF-based SLAM with Trajectory Smoothness Constraints
}

\author{Yicheng He$^{1,2}$, Guangcheng Chen$^{1}$, and Hong Zhang$^{1}$~\IEEEmembership{Fellow,~IEEE}
\thanks{$^{1}$Shenzhen Key Laboratory of Robotics and Computer Vision, Southern
University of Science and Technology, Shenzhen, China.}%
\thanks{$^{2}$Department of Electronic and Electrical Engineering, Southern University of Science and Technology, Shenzhen, China. {\tt \small heyc2023@mail.sustech.edu.cn}}%
}

\begin{document}

\maketitle
\thispagestyle{empty}
\pagestyle{empty}

\input{chapter/abstract}
\input{chapter/introduction}
\input{chapter/related_works}
\input{chapter/method}
\input{chapter/experiments}
\input{chapter/conclusions}


\bibliographystyle{IEEEtran}
\balance
\bibliography{IEEEabrv, ref}
\end{document}

%% file: chapter/abstract.tex
\begin{abstract}
The joint optimization of Neural Radiance Fields (NeRF) and camera trajectories has been widely applied in SLAM tasks due to its superior dense mapping quality and consistency. NeRF-based SLAM learns camera poses using constraints by implicit map representation. A widely observed phenomenon that results from the constraints of this form is jerky and physically unrealistic estimated camera motion, which in turn affects the map quality.
To address this deficiency of current NeRF-based SLAM, we propose in this paper TS-SLAM (TS for Trajectory Smoothness). It introduces smoothness constraints on camera trajectories by representing them with uniform cubic B-splines with continuous acceleration that guarantees smooth camera motion.
Benefiting from the differentiability and local control properties of B-splines, TS-SLAM can incrementally learn the control points end-to-end using a sliding window paradigm.
Additionally, we regularize camera trajectories by exploiting the dynamics prior to further smooth trajectories. Experimental results demonstrate that TS-SLAM achieves superior trajectory accuracy and improves mapping quality versus NeRF-based SLAM that does not employ the above smoothness constraints.
\end{abstract}

%% file: chapter/introduction.tex
\section{INTRODUCTION}
Simultaneous Localization and Mapping (SLAM) is 
widely studied for robotic systems to perform localization and scene reconstruction.
After decades of relentless research, numerous sophisticated SLAM systems have emerged.
However, the sparse maps generated by traditional SLAM systems are often inadequate for tasks such as scene understanding and path planning.
The neural radiance fields (NeRF)~\cite{mildenhall2021nerf}, a technique that utilizes multi-layer perceptrons (MLP) for the continuous representation of scenes, addresses this limitation, and it works by minimizing the color differences between the captured images and rendered images.

Recently, NeRF has been integrated into SLAM systems to reconstruct high-quality, continuous maps, and this type of SLAM is called NeRF-based SLAM and NeRF-SLAM for short in this paper.
\begin{figure}[t]
\centering
\includegraphics[width=(\textwidth-\columnsep)/2]{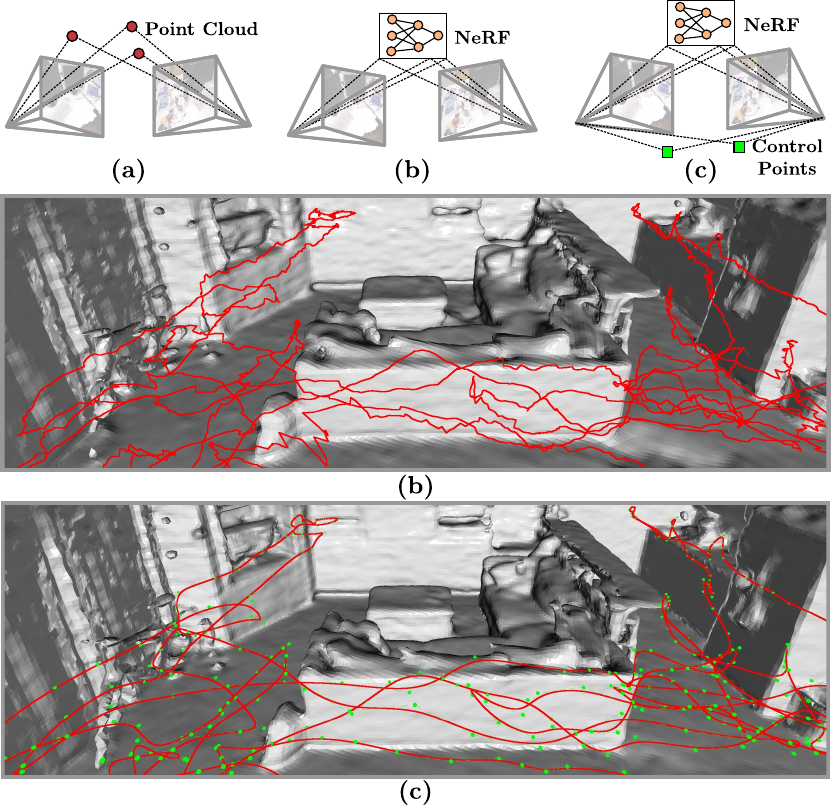}
\caption{\textbf{Camera pose constraint schemes} for (a) traditional SLAM, (b) coupled NeRF-SLAM, and (c) our method.
The estimated trajectory (red line) of current coupled NeRF-SLAM is jerky due to indirectly constrained optimization.
Our method improves trajectory accuracy and enhances reconstruction quality by introducing smoothness constraints derived from the B-spline representation of the camera trajectory.}
\label{fig:show_img}
\vspace*{-0.25in}
\end{figure}
NeRF-SLAM can be decoupled or coupled according to camera tracking strategies~\cite{tosi2024nerfs}. The former approach utilizes the camera poses computed by external trackers and images as inputs to NeRF for dense mapping~\cite{rosinol2023nerf, chung2023orbeez}.
The latter treats the camera poses as learnable parameters alongside their implicit map representation~\cite{liso2024loopy, 10610865, sucar2021imap,zhu2022nice,wang2023co,johari2023eslam,cartillier2024slaim}.
In coupled NeRF-SLAM, the camera poses and the map are unified in one differentiable computation graph,
and learned by minimizing photometric and geometric errors to achieve localization and mapping.
This simple yet efficient paradigm has been extensively studied recently, as it characterizes SLAM as a complete optimization problem that eliminates any external pre-computed information (i.e., keypoints~\cite{chung2023orbeez,10610000, campos2021orb, mao2024ngel} or
a pre-trained network~\cite{teed2024deep,teed2021droid}).
We focus on coupled NeRF-SLAM in our study due to its elegant formulation and 
performance on par with the decoupled alternative.

Unlike traditional SLAM using explicit multi-view geometric constraints,
coupled NeRF-SLAM learns camera poses constrained by implicit map representation
without explicit supervision of the underlying spatial relationship of camera poses.
These indirect constraints lead to 
camera trajectories that are globally reasonable but locally jerky and unrealistic, as shown in Fig.~\ref{fig:show_img} (b) (red curve).
Such trajectories are obviously inaccurate and can adversely affect downstream tasks such as navigation and route teach-and-repeat operation which require precise localization.
Besides, inaccurate trajectories also harm the quality of the reconstructed maps.

In this paper, we propose Trajectory Smoothness SLAM (TS-SLAM), aiming to improve the camera tracking performance of NeRF-SLAM.
TS-SLAM employs uniform cubic B-splines to represent camera trajectories, which indirectly constrains the smoothness of the trajectories, and it is able to learn the control points of the B-splines directly.
Rather than addressing each camera pose individually, TS-SLAM uses control points (green points in Fig.~\ref{fig:show_img} (c)) to constrain camera poses that are temporally close.
This representation can be regarded as a low-pass filter with supporting physical explanations~\cite{biagiotti2013online}, which prevents abrupt jumps and avoids physically irrational movements of estimated trajectories.
Besides, to stabilize the learning process, local bundle adjustment is proposed to jointly optimize the map and the control points from coarse to fine within a sliding window.
Furthermore, we use dynamics regularization to penalize predicted poses that do not conform to a dynamics prior regarding object acceleration.
Since our proposed system does not require modifications to the implicit scene representation and reconstruction loss,
its components can be seamlessly integrated into any coupled NeRF-SLAM systems.

%% file: chapter/related_works.tex
\section{RELATED WORKS}
\noindent \textbf{Neural Radiance Fields.} Neural Radiance Fields (NeRF)~\cite{mildenhall2021nerf} utilize neural networks to map 3D coordinates to their corresponding geometric and appearance information, which can then be rendered into RGB images through volumetric rendering. The neural networks are optimized by minimizing the loss between the rendered and captured images.
Subsequent works have focused on improving the quality of novel view synthesis and reducing training time. Various parameter encoding methods have been proposed, such as Tri-planes~\cite{wang2023pet}, hash encoding~\cite{muller2022instant}, Octree~\cite{yu2021plenoctrees} or voxel grid~\cite{liu2020neural}. Additionally, some works focus on the geometric reconstruction of 3D scenes. These methods propose new forms of geometric representations~\cite{guo2022neural, wang2021neus} and incorporate depth images for supervision~\cite{wang2022go, azinovic2022neural}.
To reduce the training time of the map and accurately represent detailed geometry, our proposed TS-SLAM employs hash encoding and directly predicts truncated signed distance supervised by RGBD images.

\vspace{0.15\baselineskip}
\noindent \textbf{Neural Implicit SLAM.} Due to the superior continuous representation of 3D scenes provided by NeRF, NeRF has recently been widely applied to SLAM to improve the sparse maps created by traditional SLAM~\cite{sucar2021imap,zhu2022nice,wang2023co,johari2023eslam,cartillier2024slaim}. iMAP~\cite{sucar2021imap} treats camera poses as learnable parameters for localization.
The camera poses and MLPs are jointly optimized incrementally through tracking and mapping processes.
To enhance reconstruction quality of large indoor scenes, NICE-SLAM~\cite{zhu2022nice} incorporates a multi-level feature grid and pre-trained feature decoders for scene representation.
Co-SLAM~\cite{wang2023co} adopts a joint coordinate and parametric encoding with tiny MLPs as the scene representation and trains it with global bundle adjustment.
While these works focus on improving mapping capability and often overlook camera tracking, our work introduces uniform cubic B-splines for improving pose estimation.

\vspace{0.15\baselineskip}
\noindent \textbf{SLAM with B-Splines.} Due to their useful properties, B-splines have been adopted in SLAM, leading to a category known as continuous-time SLAM. The control points of B-splines in these SLAM systems are treated as optimizable parameters.
Integrating B-splines allows SLAM systems to fuse asynchronous and high-frequency multi-sensor data without requiring timestamp alignment~\cite{yang2021asynchronous,furgale2012continuous,furgale2013unified,lv2021clins}.
B-splines also aid in modeling distortions caused by sensor motion, such as the rolling shutter effect~\cite{li2023usb, lovegrove2013spline,kerl2015dense,oth2013rolling} and motion blurs in images.
A trajectory represented by B-splines is naturally smooth and conforms to physical laws.
In this work, we consider B-splines as a learnable low-pass filter to enforce smoothness in estimated camera trajectories.

%% file: chapter/method.tex
\section{METHODOLOGY}
\begin{figure}[t]
\centering
\includegraphics[width=(\textwidth-\columnsep)/2]{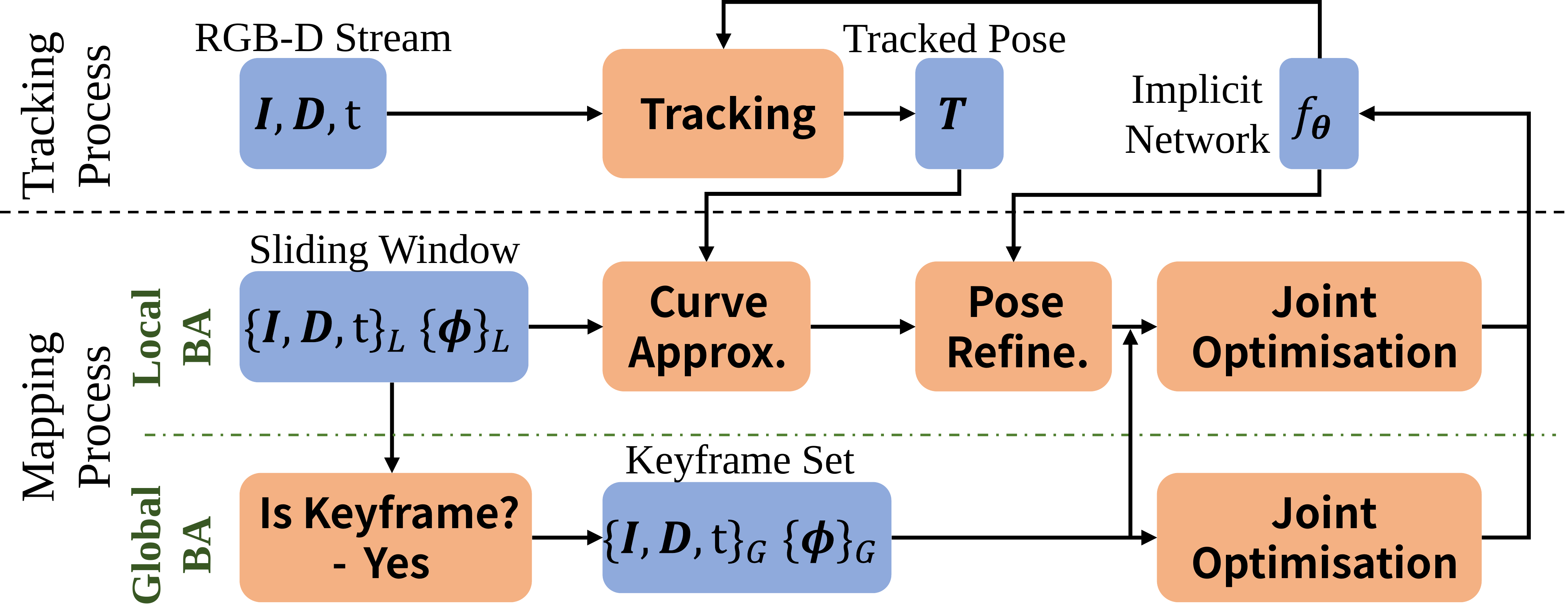}
\caption{\textbf{TS-SLAM system pipeline.} TS-SLAM system consists of two parallel threads: tracking and mapping. The mapping thread includes local and global Bundle Adjustment (BA) to optimize control points and the map.}
\label{fig:frame_work}
\vspace*{-0.25in}
\end{figure}
In this section, we present the details of TS-SLAM and Fig.~\ref{fig:frame_work} overviews how our system works. We adopt the scene representation from Co-SLAM~\cite{wang2023co} and add smoothness constraints to its optimization framework. We will introduce the representation of scene and reconstruction loss in Sec.~\ref{sec:rep_secene}, the representation of camera trajectory in Sec.~\ref{sec:rep_traj}, the dynamics regularization in Sec.~\ref{sec:dyn_reg} and the overall design of our system in Sec.~\ref{sec:sys_desp}.

\subsection{Implicit Scene Representation and Reconstruction Loss}\label{sec:rep_secene}
Given a pixel coordinate $[u,v]^T$, the camera's translation $\mathbf{t}$, rotation $\mathbf{R}$ and intrinsic matrix $\mathbf{K}$,
we employ the neural scene representation $f_{\theta}$ defined in Co-SLAM~\cite{wang2023co}. This representation maps world coordinates $\mathbf{x}$ to color $\mathbf{c}$ and truncated signed distance function (TSDF) $s$ values:
\begin{gather}
f_{\theta}(\mathbf{x}_i) \to (\mathbf{c}_i, s_i),\\
\mathbf{x}_i = \mathbf{t} + d_i\mathbf{R}\mathbf{K}^{-1}[u,v,1]^T,~~i \in \{1, ..., L\}\label{eq:cal_ray},
\end{gather}
where $f_{\theta}$ is parameterized by a fully connected neural network and a multi-resolution hash-based feature grid. Using the network outputs $\{(\mathbf{c}_i, s_i)\}$, we render the color $\hat{\mathbf{c}}$ and depth $\hat{d}$ at $[u,v]^T$ through volume rendering:
\begin{equation}
\hat{\mathbf{c}} = \frac{1}{\sum_{i=1}^{L}w_i}\sum_{i=1}^{L}w_i\mathbf{c}_i,~~\hat{d} = \frac{1}{\sum_{i=1}^{L} w_i}\sum_{i=1}^Lw_id_i,
\end{equation}
where $\{w_i = \sigma(s_i/tr)\sigma(-s_i/tr)\}$ are the weights along the ray which are multiplication of two Sigmoid functions related to the TSDF values and truncation distance $tr$. In TS-SLAM, the scene representation $f_{\theta}$ and camera poses $[\mathbf{R}~\mathbf{t}]$ can be learned end-to-end through the loss function:
\begin{equation}
\mathcal{L}_{NeRF} = \mathcal{L}_{rgb} + \mathcal{L}_{d} + \mathcal{L}_{sdf} + \mathcal{L}_{fs} + \mathcal{L}_{smooth}.
\label{eq:lossfuc}
\end{equation}
$\mathcal{L}_{rgb}$ and $\mathcal{L}_{d}$ represent the pixel-to-pixel loss functions for the captured RGB and depth images compared to the rendered RGB and depth images, respectively. 
$\mathcal{L}_{sdf}$ utilize depth images to supervise SDF values and the free-space loss $\mathcal{L}_{fs}$ forces the SDF prediction which is far from the surface to be the truncated distance.
$\mathcal{L}_{smooth}$ prevents artifacts of the reconstructions caused by hash collisions. Once $f_{\theta}$ converges, the construction of the map is complete. For more details, please refer to \cite{wang2023co}.
Additionally, a dynamics regularization term is added to $\mathcal{L}_{NeRF}$, which will be discussed in detail in Sec.~\ref{sec:dyn_reg}.

\begin{figure}[t]
\centering
\includegraphics[width=(\textwidth-\columnsep)/2]{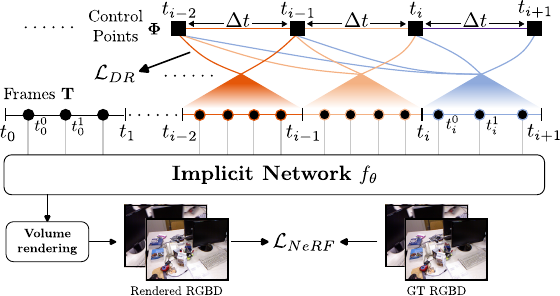}
\caption{\textbf{End-to-end learning of control points.} Four control points influence the pose at a certain moment on the curve, indirectly introducing smoothness constraints among camera poses that are temporally close. The control points are learned end-to-end by minimizing $\mathcal{L}_{NeRF}$ and $\mathcal{L}_{DR}$.}
\label{fig:factor-graph}
\vspace*{-0.25in}
\end{figure}

\subsection{B-splines Representation of Trajectory}\label{sec:rep_traj}
At the core of our approach lies the B-spline trajectory representation. As cubic splines with $C^2$ continuity imply smoothness, B-splines are locally controlled, meaning any point on the spline depends on a few control points localized nearby. 
These properties make B-splines well-suited for smoothing SLAM trajectories where incrementally estimating camera poses is required.
Therefore, we parameterize the trajectory using uniform cubic B-splines.

Uniform cubic B-splines assume that control points are placed at times $\{t_i\}$, with a constant interval $\Delta t$ between them.
The pose at any given time $t$ on cubic B-splines is influenced by four control points.
We define these control points as the set
$\{\mathbf{\Phi}_k=[\mathbf{R}_k~\mathbf{t}_k]\ |\ k=i-2,i-1,i,i+1\}$
for the interval $t \in [t_i ,t_{i+1})$.
To simplify calculations, we use a uniform time scale $s(t) := (t - t_0) / \Delta t$, converting the times of the control points $\{t_i\}$ to uniform indices $\{s_i\}$. Given a time index $s_i \leq s(t) < s_{i+1}$, we define $u(t) = s(t) - s_i$. Using this time formulation and based on the matrix representation for the De Boor-Cox formula~\cite{cox1972numerical,de1972calculating}, we can write the cumulative form of uniform cubic B-splines~\cite{qin1998general} for translation $\mathbf{t}\in \mathbb{R}^3$ and rotation $\mathbf{R} \in SO(3)$ in $t \in [t_i, t_{i+1})$:
\begin{align}
\mathbf{t}(t) &= \mathbf{t}_{i-2} + \sum_{j=1}^{3}\mathbf{B}_j(u)(\mathbf{t}_{i+j-2} - \mathbf{t}_{i+j-3}), \label{eq:bspline_p}\\
\mathbf{R}(t) &= \mathbf{R}_{i-2} \prod_{j=1}^{3} \operatorname{Exp}(\mathbf{B}_j(u) \mathbf{d}_j), \label{eq:bspline_r}\\
\mathbf{B}(u)& = \mathbf{C}\begin{bmatrix}
1\\
u\\
u^2\\
u^3
\end{bmatrix}, \mathbf{C} = \frac{1}{6}\begin{bmatrix}
6 & 0 & 0 & 0 \\
5 & 3 & -3 & 1 \\
1 & 3 & 3 & -2 \\
0 & 0 & 0 & 1 \\
\end{bmatrix},\nonumber
\end{align}
with the generalized difference vector $\mathbf{d}_j$
\begin{equation}
\mathbf{d}_j = \operatorname{Log}(\mathbf{R}^{-1}_{i+j-3}\mathbf{R}_{i+j-2}) \in \mathfrak{se}3.
\end{equation}
$\mathbf{T}(t)=[\mathbf{R}(t)~\mathbf{t}(t)]$ is the pose along the spline curve at time $t$, which is used in Eq.~\eqref{eq:cal_ray}.
$\mathbf{B}_j(u)$ is the $j$-th element of $\mathbf{B}(u)$ (zero-based numbering). Note that $\mathbf{B}(u)$ is constant for each $t$.

Since camera poses are the weighted compositions of their neighbour control points according to Eq.~\eqref{eq:bspline_p} and Eq.~\eqref{eq:bspline_r}, they are differentiable w.r.t. the control points. 
This allows TS-SLAM to jointly optimize the map and control points in an end-to-end manner.
Each control point influences frames that are temporally close on the trajectory, as shown in Fig.~\ref{fig:factor-graph}.
Furthermore, poses interpolated using cubic B-splines are naturally $C^2$ continuous.
Based on these characteristics, B-splines introduce geometric smoothness constraints among camera poses that are temporally close.

\subsection{Dynamics Regularization}\label{sec:dyn_reg}
The acceleration of camera motion has an upper bound due to the limited torque and force available in the system. We introduce a dynamics regularization for the trajectories to integrate the physical priors.
Thanks to the $C^2$ continuity of the B-splines, we can calculate higher-order derivatives at any point of the trajectories, such as acceleration and angular acceleration. 
The acceleration $\mathbf{a}$ in $t \in [t_i, t_{i+1})$ is given by:
\begin{gather}
\mathbf{a}(t) = \sum_{j=1}^{3}\mathbf{\ddot{B}}_j(u)(\mathbf{t}_{i+j-2} - \mathbf{t}_{i+j-3}),\label{eq_jeck}\\
\mathbf{\ddot{B}}(u) = \frac{1}{\Delta t^2}\mathbf{C}\begin{bmatrix}
0&
0&
2&
6u
\end{bmatrix}^T.\nonumber
\end{gather}
For angular acceleration $\boldsymbol{\dot{\omega}}$, we compute it recursively to achieve faster computation speeds~\cite{sommer2020efficient}.
With $\mathbf{A}_j(u) = \operatorname{Exp}(\mathbf{B}_j(u)\cdot\mathbf{d}_j)$ and $\boldsymbol{\omega}^{(1)} = \boldsymbol{\dot{\omega}}^{(1)} = \mathbf{0} \in \mathbb{R}^3$,
we omit $t$ for brevity and calculate $\boldsymbol{\dot{\omega}}$ as follows:
\begin{align}
&\boldsymbol{\dot{\omega}}^{(j)} = \mathbf{\dot{B}}_{j-1}\boldsymbol{\omega}^{(j)} \times \mathbf{d}_{j-1} + \mathbf{A}^T_{j-1}\boldsymbol{\dot{\omega}}^{(j-1)} + \mathbf{\ddot{B}}_{j-1}\mathbf{d}_{j-1},\label{eq_rot_jeck}\\
&\boldsymbol{\omega}^{(j)} = \mathbf{A}^T_{j-1}\boldsymbol{\omega}^{(j-1)} + \mathbf{\dot{B}}_{j-1}\mathbf{d}_{j-1},\nonumber
\end{align}
where
\begin{equation}
\mathbf{\dot{B}}(u) = \frac{1}{\Delta t}\mathbf{C}\begin{bmatrix}
0&
1&
2u&
3u^2
\end{bmatrix}^T.\nonumber
\end{equation}
The cubic B-splines require three iterations of Eq.~\eqref{eq_rot_jeck}, resulting in $\boldsymbol{\dot{\omega}}^{(4)}$, which is the rotational acceleration of cubic B-splines. By constraining $\|\mathbf{a}\|$ and $\|\boldsymbol{\dot{\omega}}\|$, we can integrate physical priors into the system.
By Combining dynamics regularization $\mathcal{L}_{DR}$ with Eq.~\eqref{eq:lossfuc}, the total loss function $\mathcal{L}_{total}$ is given by:
\begin{align}
\mathcal{L}_{NeRF} - \sum_{i=1}^{K}(\lambda_1\ln(1-\frac{\|\mathbf{a}_i\|}{\|\mathbf{a}\|_{max}}) + \lambda_2\ln(1-\frac{\|\boldsymbol{\dot{\omega}}_i\|}{\|\boldsymbol{\dot{\omega}}\|_{max}})),\label{eq:loss_ba}
\end{align}
where $\lambda_1$ and $\lambda_2$ are positive weight coefficients, $\|\mathbf{a}\|_{max}$ and $\|\boldsymbol{\dot{\omega}}\|_{max}$ are the upper bound of $\|\mathbf{a}\|$ and $\|\boldsymbol{\dot{\omega}}\|$ depending on the system's control strategy. This regularization term uniformly samples $K$ timestamps and calculates the corresponding $\mathbf{a}$ and $\boldsymbol{\dot{\omega}}$ using Eq.~\eqref{eq_jeck} and Eq.~\eqref{eq_rot_jeck}, subsequently penalizing values that approach the upper bound. 
Representing the trajectory through B-splines guarantees continuous acceleration, while dynamics regularization further constrains the magnitude of the acceleration.
In addition to incorporating physical priors for a more physically realistic trajectory, this regularization term can be applied at any point along the trajectory, allowing the total loss function to affect not only moments with image input. This increases the sampling rate of the loss function on the camera trajectory, effectively reducing excessive oscillations in the estimated motion.

\subsection{System Description}\label{sec:sys_desp}
\subsubsection{Overview}
As shown in Fig.~\ref{fig:frame_work}, TS-SLAM consists of two parallel threads: tracking and mapping, using an RGBD stream with the timestamps 
for each frame as input. TS-SLAM freezes the map in the tracking thread and optimizes the camera poses individually.
The mapping thread is divided into local and global bundle adjustment (BA). Lobal BA uses the camera poses learned by the tracking thread to initialize control points and jointly optimizes the control points in a sliding window and the map. In global BA, the map and the control points corresponding to keyframes are optimized simultaneously.

\vspace{0.15\baselineskip}
\subsubsection{Tracking Thread}
The system estimates the transformation from camera to world coordinates for each frame.
When a new frame is received, we initialize the camera pose using a constant velocity motion model.
Then, we uniformly sample pixels on the current frame and iteratively optimize the pose of the current frame by minimizing $\mathcal{L}_{total}$ in Eq. \eqref{eq:loss_ba} with $\lambda_1 = \lambda_2 = 0$.

\vspace{0.15\baselineskip}
\subsubsection{Local Bundle Adjustment}\label{sec:lba}
\begin{figure}[t]
\centering
\includegraphics[width=(\textwidth-\columnsep)/2]{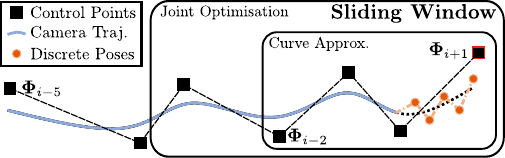}
\caption{\textbf{Local bundle adjustment.} 
The sliding window contains the latest $M$ control points and the RGBD observations within the interval $[t_{i+1-M}, t_{i+1})$. 
The red-framed square represents the newly added control point that needs to be initialized. The discrete poses are the camera poses output by the tracking thread.}
\label{fig:sliding_window}
\vspace*{-0.25in}
\end{figure}

For every interval $\Delta t$, TS-SLAM runs a mapping thread consisting of local BA and global BA. Local BA optimizes the scene representation and the control points in a sliding window to achieve a smooth camera trajectory. The local control property of B-splines enables the implementation of local BA using a sliding window, as illustrated in Fig.~\ref{fig:sliding_window}. Specifically, the sliding window contains the most recent $M$ control points and has a step size of one control point.

The local BA uses discrete camera poses to initialize each newly added control point through curve approximation, accelerating the convergence of the control points.
For the image timestamps $\{t_i^k\}_{k=1}^{N}$ in interval $[t_{i}, t_{i+1})$, the tracking process outputs corresponding camera poses $\{\mathbf{T}_{i}^k\}_{k=1}^N$, as shown by the red points in Fig.~\ref{fig:sliding_window}, where $N$ depends on the camera frame rate and the time interval $\Delta t$ of the B-spline curve.
We interpolate the corresponding camera poses $\{\hat{\mathbf{T}}(t_i^{k})\}_{k=1}^N$ using the latest four control points $\{ \mathbf{\Phi}_{i-2} ,\mathbf{\Phi}_{i-1} ,\mathbf{\Phi}_{i} ,\mathbf{\Phi}_{i+1} \}$ by Eq.~\eqref{eq:bspline_p} and Eq.~\eqref{eq:bspline_r}. 
The control point $\mathbf{\Phi}_{i+1}$ at the newly added time $t_{i+1}$ is initialized by curve approximation:
\begin{equation}
\operatorname*{arg\min}_{\mathbf{\Phi}_{i+1}}
\sum_{k=1}^{N}(||\operatorname{Log}((\mathbf{R}_{i}^k)^T\hat{\mathbf{R}}(t^k_i))|| + ||\mathbf{t}_{i}^k - \hat{\mathbf{t}}(t^k_i)||).
\end{equation}
Note that we need to initialize all four control points instead of only the last one in the first time interval $[t_0, t_1)$ for system initialization.

After curve approximation, we further refine the control points to enhance their consistency with the map. Specifically, the control points are used to compute the poses of all frames within the sliding window which are then further refined by minimizing $\mathcal{L}_{total}$ in Eq.~\eqref{eq:loss_ba} while the map $f_\theta$ is frozen. For the dynamics regularization, the acceleration $\mathbf{a}$ and rotational acceleration $\boldsymbol{\dot{\omega}}$ are uniformly sampled within the sliding window. 
Finally, we select a keyframe every five frames and jointly optimize the map and control points.

Local BA can be considered an optimization process for control points from coarse to fine. This improves the stability of control point optimization and reduces the errors introduced by newly added control points to subsequent global BA.

\vspace{0.15\baselineskip}
\subsubsection{Global Bundle Adjustment}
Global BA jointly optimizes the map and the control points corresponding to all keyframes outside the sliding window. 
The loss function is the same as that of local BA, except that the dynamics regularization is applied outside the sliding window. Global BA prevents catastrophic forgetting of NeRF and enhances the consistency between the camera trajectory and the map.

%% file: chapter/experiments.tex
\section{EXPERIMENTS}
\begin{figure*}[ht]
\centering
\includegraphics[width=0.975\textwidth]{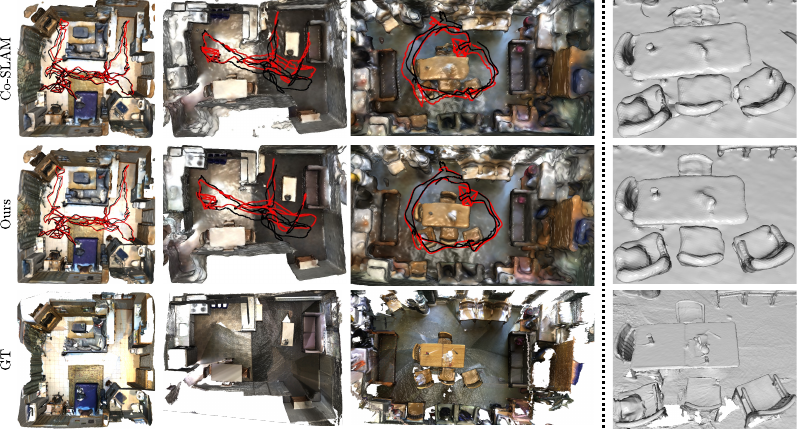}
\caption{\textbf{Qualitative results of TS-SLAM on the ScanNet dataset.} The ground truth trajectory is shown in black, and the estimated trajectory is shown in red. Our method achieves more accurate camera tracking results than the baselines and improves mapping quality (right column).}
\label{fig:scannet_show}
\vspace*{-0.1in}
\end{figure*}
\begin{table*}[t]
\caption{\textbf{ATE RMSE (cm) and RPE RMSE (cm) results averaged over three runs on ScanNet.} We align estimated and ground truth trajectories using Horn’s method [31] before evaluating the ATE. TS-SLAM outperforms baselines.}
\renewcommand{\arraystretch}{1.3}
\begin{center}
\begin{tabular}{cccccccc|ccccccc}
\toprule
\multicolumn{1}{c}{}                                                    & \multicolumn{7}{c|}{ATE RMSE (cm)}                                                                            & \multicolumn{7}{c}{RPE RMSE (cm)}                                                                           \\ \cmidrule{2-15}
Scene ID                                                                & 0000          & 0059          & 0106          & 0169          & 0181          & 0207          & Avg.          & 0000          & 0059          & 0106          & 0169          & 0181         & 0207         & Avg.          \\ \midrule
\multicolumn{1}{c|}{iMAP~\cite{sucar2021imap}}    & 55.95         & 18.80         & 13.92         & 20.78         & 35.40         & 11.91         & 26.12         & 1.70          & 1.66          & 1.17          & 1.97          & 1.71         & 1.55         & 1.62          \\
\multicolumn{1}{c|}{NICE-SLAM~\cite{zhu2022nice}} & 12.06          & 13.25         & 8.21          & 14.4         & 13.05         & \textbf{5.98}          & 11.15          & 1.01          & 1.62          & 0.91          & 1.11          & 1.34         & 1.03         & 1.17          \\
\multicolumn{1}{c|}{Co-SLAM~\cite{wang2023co}}    & 7.13          & 11.14         & 9.36          & 5.90          & 11.81         & 7.14          & 8.75          & 1.49          & 2.04          & 1.29          & 1.41          & 1.82         & 1.41         & 1.58          \\
\multicolumn{1}{c|}{Ours}                                               & \textbf{5.89} & \textbf{8.20} & \textbf{7.79} & \textbf{5.73} & \textbf{9.80} & 6.40 & \textbf{7.30} & \textbf{0.82} & \textbf{1.19} & \textbf{0.71} & \textbf{0.84} & \textbf{1.18} & \textbf{0.90} & \textbf{0.94} \\ \bottomrule
\end{tabular}
\end{center}
\label{tab:scannet_track}
\vspace*{-0.2in}
\end{table*}
\subsection{Experimental Setup}
\subsubsection{Datasets}
We evaluate TS-SLAM on three datasets. Similar to previous work, we evaluate the tracking performance on three scenes of the TUM RGBD dataset~\cite{sturm2012benchmark} with their poses provided by the motion capture system and six real scenes of the ScanNet dataset~\cite{dai2017scannet} with their ground truth pose derived using BundleFusion~\cite{dai2017bundlefusion}. Following Co-SLAM~\cite{wang2023co}, we compare reconstruction performance on seven synthetic scenes from NeuralRGBD~\cite{azinovic2022neural}, which simulates noisy depth images. Due to the absence of timestamps for individual frames in ScanNet~\cite{dai2017scannet} and the synthetic dataset~\cite{azinovic2022neural}, we assumed a constant frame rate of 30 FPS.

\vspace{0.15\baselineskip}
\subsubsection{Metrics}
We assess camera tracking using ATE RMSE ($cm$) and RPE RMSE ($cm$)~\cite{sturm2012benchmark}. Consistent with previous work, we align the estimated and ground truth trajectories using Horn's method~\cite{horn1987closed} before evaluating ATE. RPE measures the trajectory accuracy locally over a fixed interval set to 1 for analyzing the drift per frame. Compared with ATE, which targets global accuracy, RPE focuses on local discrepancies and better examines trajectory smoothness. For reconstruction, we employ Depth L1 ($cm$), Accuracy ($cm$), Completion ($cm$), and Completion Ratio ($\%$) with a 5 cm threshold, using the same mesh culling strategy as Co-SLAM~\cite{wang2023co} before evaluation.

\vspace{0.15\baselineskip}
\subsubsection{Baselines}
We considers iMAP~\cite{sucar2021imap}, NICE-SLAM~\cite{zhu2022nice}, and Co-SLAM~\cite{wang2023co} as our baselines. Since TS-SLAM adopts the map representation of Co-SLAM~\cite{wang2023co}, Co-SLAM~\cite{wang2023co} serves as our primary baseline.

\vspace{0.15\baselineskip}
\subsubsection{Implementation Details}
We run TS-SLAM on a single NVIDIA RTX 3090 GPU. For  synthetic dataset~\cite{azinovic2022neural} and ScanNet~\cite{dai2017scannet}, we set $\Delta t=0.3s$, and for TUM~\cite{sturm2012benchmark}, $\Delta t=0.25s$. Weight coefficients $\lambda_1 = \lambda_2 = 0.1$. $\|\mathbf{a}\|_{max}$ and $\|\mathbf{\dot{\omega}}\|_{max}$ are set to 5 and 5, respectively. All mapping-related parameters are kept consistent with those used in Co-SLAM~\cite{sucar2021imap}.

\begin{table}[t]
\renewcommand{\arraystretch}{1.3}
\caption{\textbf{Tracking performance on TUM RGBD dataset.} Our method outperforms the baselines, particularly in local trajectory precision, indicated by the RPE RMSE.}
\renewcommand\arraystretch{1.20}
  \renewcommand\tabcolsep{5pt}
	\normalsize
\begin{center}
\resizebox{\linewidth}{!}{
\begin{tabular}{cccc|ccc}
\toprule
\multirow{2}{*}{}              & \multicolumn{3}{c|}{ATE RMSE (cm)}         & \multicolumn{3}{c}{RPE RMSE (cm)}             \\  \cmidrule{2-7}
                               & fr1/desk    & fr2/xyz     & fr3/office  & fr1/desk     & fr2/xyz      & fr3/office   \\ \midrule
\multicolumn{1}{c|}{iMAP~\cite{sucar2021imap}}      & 4.9          & 2.0          & 5.8          & 2.32          & 1.23      & 1.46             \\
\multicolumn{1}{c|}{NICE-SLAM~\cite{zhu2022nice}} & 2.7          & 1.8          & 3.0          & 1.03          & 0.82         & 1.30             \\
\multicolumn{1}{c|}{Co-SLAM~\cite{wang2023co}}   & 2.5          & 1.7          & 2.4          & 1.23          & 0.75          & 0.82          \\
\multicolumn{1}{c|}{Ours}      & \textbf{2.2} & \textbf{1.6} & \textbf{2.4} & \textbf{0.31} & \textbf{0.11} & \textbf{0.18} \\ \bottomrule
\end{tabular}}
\end{center}
\label{tab:tum_track}
\vspace*{-0.15in}
\end{table}

\begin{table}[!t]
\renewcommand{\arraystretch}{1.3}
\caption{\textbf{Reconstruction quality on NeuralRGBD dataset averaged over three runs.} TS-SLAM achieves more accurate camera tracking, thereby improving reconstruction quality.}
\begin{center}
\resizebox{\linewidth}{!}{
\begin{tabular}{ccccc}
\toprule
          & Depth L1 (cm)$\downarrow$ & Acc. (cm)$\downarrow$     & Comp. (cm)$\downarrow$    & Comp. Ratio$\uparrow$   \\ \midrule
iMAP~\cite{sucar2021imap}      & 43.96         & 18.30         & 26.41         & 20.73         \\
NICE-SLAM~\cite{zhu2022nice} & 6.32          & 5.96          & 5.30          & 77.46         \\
Co-SLAM~\cite{wang2023co}   & 3.02          & 2.95          & 2.96          & 86.88         \\
Ours      & \textbf{2.87} & \textbf{2.79} & \textbf{2.87} & \textbf{87.02} \\ \bottomrule
\end{tabular}}
\end{center}
\label{tab:syn}
\vspace*{-0.25in}
\end{table}
\subsection{Tracking and Reconstruction Performance}
\subsubsection{Camera Tracking}
\begin{figure}[t]
\centering
\includegraphics[width=(\textwidth-\columnsep)/2]{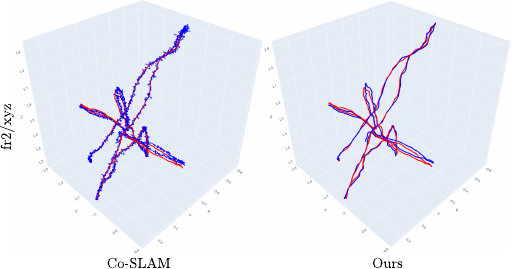}
\caption{\textbf{Qualitative results of camera tracking on the TUM dataset.} The ground truth trajectory is shown in red, and the aligned estimated trajectory is in blue. Our method effectively guarantees smoothness of camera motion.}
\label{fig:tum_show}
\vspace*{-0.1in}
\end{figure}

We evaluate camera tracking on six room-scale sequences from ScanNet~\cite{dai2017scannet} and three real scenes from TUM RGBD~\cite{sturm2012benchmark}. Trajectory accuracy is assessed using ATE for global accuracy and RPE for local errors. The results are shown in Tab.~\ref{tab:scannet_track} and Tab.~\ref{tab:tum_track}. Our method demonstrates a significant improvement in localization accuracy, especially in terms of local accuracy, as reflected by the RPE. Compared to Co-SLAM~\cite{wang2023co}, TS-SLAM achieves reductions of 78.5\% and 40.5\% in RPE on TUM~\cite{sturm2012benchmark} and ScanNet~\cite{dai2017scannet}, respectively.
The smoothness constraints of TS-SLAM significantly avoid trajectory discontinuity, as shown by the aligned trajectories in Fig.~\ref{fig:tum_show}, and improve global accuracy as shown in Fig.~\ref{fig:scannet_show}.
TS-SLAM connects camera poses that are close in time, which can be understood as treating each trajectory segment as a whole and optimizing it through control points, and thus can reduce trajectory drift, as demonstrated by experimental results.

\vspace{0.15\baselineskip}
\subsubsection{Scene Reconstruction}
We evaluate the reconstruction quality on the seven synthetic scenes from NeuralRGBD~\cite{azinovic2022neural}. Unlike other synthetic datasets, it simulates noise in the depth sensor data. TS-SLAM directly adopts map representation of Co-SLAM without any modifications. Our method can clearly improve reconstruction quality, as shown in Tab.~\ref{tab:syn}. The trajectory smoothness constraints lead to more accurate camera tracking and therefore more accurate reconstruction results, as shown in the right column of Fig.~\ref{fig:scannet_show}.
\begin{table}[t]
\renewcommand{\arraystretch}{1.3}
\caption{\textbf{Runtime comparison across baseline}. Runtime is reported in $ms/iter \times \#iter$.
TS-SLAM runs the mapping thread approximately every 10 frames on ScanNet and every 8 frames on TUM.}
\begin{center}
\resizebox{\linewidth}{!}{
\begin{tabular}{ccccccc}
\toprule
 & \multirow{2}{*}{Method} & \multirow{2}{*}{Track.} & \multicolumn{2}{c}{LBA} & \multirow{2}{*}{GBA} & \multirow{2}{*}{FPS} \\ \cline{4-5}
                         &                         &                              & Init.   & Others       &                                 &                      \\ \hline
\multirow{3}{*}{\rotatebox{90}{ScanNet}} & NICE-SLAM~\cite{zhu2022nice}               & 12.3$\times$50               & -               & -               & 125.3$\times$60                        & 0.68                 \\
                         & Co-SLAM~\cite{wang2023co}                 & 11.2$\times$20                & -               & -               & 37.2$\times$10                  & 4.4                  \\
                         & Ours                    & 11.2$\times$20                & 23.5$\times$20  & 63.1$\times$10  & 65.5$\times$10                  & 4.4                  \\ \midrule
\multirow{3}{*}{\rotatebox{90}{TUM}}     & NICE-SLAM               & 47.1$\times$200              & -               & -               & 189.2$\times$60                 & 0.08                 \\
                         & Co-SLAM                 & 10.6$\times$20                & -               & -               & 36.0$\times$20                  & 4.7                  \\
                         & Ours                    & 10.6$\times$20                & 23.5$\times$20  & 62.2$\times$10  & 52.5$\times$20                  & 3.6                 \\ \bottomrule
\end{tabular}}
\end{center}
\label{tab:runtime}
\vspace*{-0.28in}
\end{table}
\subsection{Runtime Analysis}
Tab.~\ref{tab:runtime} reports the runtime of each process for different methods on the ScanNet~\cite{dai2017scannet} and TUM~\cite{sturm2012benchmark} datasets. The runtime is expressed as the time per iteration multiplied by the number of iterations. Co-SLAM performs global BA every five frames, while our method runs mapping thread approximately every 10 frames on ScanNet and every 8 frames on TUM.
The introduction of local BA and Pytorch's automatic differentiation for B-splines increase the computational load.
However, compared to Co-SLAM, our mapping thread runs less frequently, resulting in only a slight drop in FPS on TUM dataset.

\begin{table}[t]
\renewcommand{\arraystretch}{1.3}
\caption{
\textbf{Ablation studies on ScanNet Dataset.} We test the impact of B-spline represented trajectory (B-S.), local bundle adjustment (LBA), dynamics regularization term (DR) and time interval ($\Delta t$). The metrics represent the average RMSE (cm) across all six scenes.}
\begin{center}
\begin{tabular}{ccccccc}
\toprule
 Variations  & B-S.         & LBA          & DR      & $\Delta t$ (s) & ATE      & RPE      \\ \midrule
w/o B-S.   &              & $\checkmark$ &              & -              & 8.63          & 1.52          \\
w/o LBA   & $\checkmark$ &              &              & 0.30            & -             & -             \\
w/o DR    & $\checkmark$ & $\checkmark$ &              & 0.30            & 7.74          & 0.98          \\ \hline
${\Delta_1}$ & $\checkmark$ & $\checkmark$ & $\checkmark$ & 0.16           & 8.96          & 0.97          \\
${\Delta_2}$ & $\checkmark$ & $\checkmark$ & $\checkmark$ & 0.40           & 7.64          & 0.96          \\
${\Delta_3}$ & $\checkmark$ & $\checkmark$ & $\checkmark$ & 0.56           & 13.75         & 1.05          \\ \hline
${\Delta}^{*}$             & $\checkmark$ & $\checkmark$ & $\checkmark$ & 0.30            & \textbf{7.30} & \textbf{0.94} \\ \bottomrule
\end{tabular}
\end{center}
\label{tab:ablation}
\vspace*{-0.26in}
\end{table}

\subsection{Ablation Studies}

\subsubsection{\textbf{Effect of B-spline represented trajectory}} 
The results from the first and third rows in Tab.~\ref{tab:ablation} indicate that introducing smoothness constraints through B-splines led to 10.31\% and 35.5\% reductions in ATE and RPE, respectively. The local BA in w/o BS. removes curve approximation and optimizes the camera poses in the remaining steps.

\vspace{0.15\baselineskip}
\subsubsection{\textbf{Effect of local bundle adjustment}} When local BA is removed (w/o LBA), the system fails to run.
Without the coarse-to-fine learning strategy of local BA, the joint optimization of the map and the control points introduces unacceptable reconstruction errors, causing the system to fail. 

\vspace{0.15\baselineskip}
\subsubsection{\textbf{Effect of dynamics regularization}} Dynamics regularization introduces dynamics priors during optimization.
The results from the third and last rows in Tab.~\ref{tab:ablation} demonstrate that the dynamics regularization provides a certain degree of improvement in trajectory accuracy.

\vspace{0.15\baselineskip}
\subsubsection{\textbf{Effect of time interval}}
The fourth to seventh rows in Tab.~\ref{tab:ablation} show the experimental results of TS-SLAM with different $\Delta t$.
Empirically, TS-SLAM achieves optimal performance when $\Delta t$ approximates twice the keyframe interval ($\Delta^{*}$), ensuring sufficient frames for learning control points and maintaining an appropriate mapping frequency.

%% file: chapter/conclusions.tex
\section{CONCLUSIONS}
In this paper, we propose TS-SLAM, a dense visual SLAM approach that improves camera tracking accuracy and indirectly enhances map quality by introducing smoothness constraints through a B-spline representation of the camera trajectory. 
By leveraging the differentiability of the B-splines, TS-SLAM learns the control points end-to-end.
Additionally, we design a dynamics regularization term to penalize excessive acceleration of the camera motion, leading to a more physically realistic trajectory.
To stabilize the learning process, we utilize the local support property of B-splines and design a local bundle adjustment that optimizes the control points from coarse to fine.
Extensive experiments demonstrate that trajectory smoothness constraints can effectively improve existing NeRF-SLAM methods.

\textbf{Limitations}.
Our method uses uniform cubic B-splines to represent the trajectory, which requires predetermining the fixed time intervals and spline order.
Future work will explore how to adjust these hyperparameters adaptively.